\documentclass[runningheads]{llncs}

\usepackage{amsmath}
\usepackage{booktabs}
\usepackage{float}
\usepackage{subcaption}
\usepackage{pgfplots}
\usepackage{graphicx}
\usepackage{color, colortbl}
\usepackage[utf8]{inputenc}

\definecolor{bblue}{HTML}{4F81BD}
\definecolor{rred}{HTML}{C0504D}
\definecolor{ggreen}{HTML}{9BBB59}
\definecolor{ppurple}{HTML}{9F4C7C}
\definecolor{amber}{rgb}{1.0, 0.75, 0.0}
\definecolor{Gray}{gray}{0.8}

\newcommand\floor[1]{\lfloor#1\rfloor}

\newcommand{\rpm}{\sbox0{$1$}\sbox2{$\scriptstyle\pm$}
  \raise\dimexpr(\ht0-\ht2)/2\relax\box2 }
\begin{document}
\title{Low-Precision Floating-Point Schemes\\ for Neural Network Training}
%
%
\author{Marc Ortiz\inst{1,2} \and Adrián Cristal\inst{1} \and Eduard Ayguadé\inst{1,2} \and Marc Casas\inst{1}}
\authorrunning{M.Ortiz et al.}
%
\institute{Barcelona Supercomputing Center, Barcelona 08034, Spain\\
\email{\{marc.ortiz , adrian.cristal, eduard.ayguade, marc.casas\}@bsc.es}
\and
Universitat Politécnica de Catalunya, Barcelona 08034, Spain 
\email{\{eduard\}@ac.upc.edu}
}
\maketitle              
\begin{abstract}
The use of low-precision fixed-point arithmetic along with stochastic rounding has been proposed as a promising alternative to the commonly used 32-bit floating point arithmetic to enhance training neural networks training in terms of performance and energy efficiency. In the first part of this paper, the behaviour of the 12-bit fixed-point arithmetic when training a convolutional neural network with the CIFAR-10 dataset is analysed, showing that such arithmetic is not the most appropriate for the training phase. After that, the paper presents and evaluates, under the same conditions, alternative low-precision arithmetics, starting with the 12-bit floating-point arithmetic. These two representations are then leveraged using local scaling in order to increase accuracy and get closer to the baseline 32-bit floating-point arithmetic. Finally, the paper introduces a simplified model in which both the outputs and the gradients of the neural networks are constrained to power--of--two values, just using 7 bits for their representation. The evaluation demonstrates a minimal loss in accuracy for the proposed Power--of--Two neural network, avoiding the use of multiplications and divisions and thereby, significantly reducing the training time as well as the energy consumption and memory requirements during the training and inference phases.\\

\end{abstract}
%
%
%

\section{Introduction}
Over the last few years, neural networks have been applied to a large variety of problems, including image description, self-driving cars, speech recognition, generation of content or even in art. 
In many of these fields, neural networks show accuracies that meet and sometimes exceed human-level performance \cite{microsoft_imagenet,microsoft_speech}.
Their success relay on the huge increase in computational resources that we have seen during the last years, allowing to increase the complexity of the neural network models as well as to use much larger data sets for training. 
While this increase in the complexity of neural models is a key-success \cite{resnet,deepConv}, it elevates the computational and energy/power requirements during the training phase, limiting their utilisation low-power environments. 
To address this, different kinds of solutions have already been proposed, such as the use of approximate computing or/and the use of high-throughput architectures (e.g. GPU to accelerate the processing of complex neural models or reconfigurable devices like FPGAs~\cite{fpga} to mitigate power budget restrictions).

This paper focuses on the use of reduced-precision techniques, which is preceded by a large body of previous work.
Indeed, there is a considerable number of recent studies adopting alternative low-precision arithmetics and data formats for the training and the inference of neural networks~\cite{fixed1,fixed3,fixed2,fixed4,fixed5}. 
All solutions show a substantial improvement in hardware footprint, power consumption, speed and memory requirements by suggesting the replacement of the commonly used 32-bit floating-point arithmetic by a low-precision fixed-point approach.
Results show that the combination of limited-precision fixed-point arithmetics with techniques like stochastic rounding makes possible for network models to operate with little or no accuracy degradations.

Despite these satisfactory results, this paper first shows that, in the scenario of training a Convolutional Neural Network with the CIFAR-10 image dataset, the fixed-point arithmetic is not the most appropriate one.
In this paper, we analyse the obstacles faced when training a neural network with the limited-precision fixed-point arithmetic and we propose and evaluate several alternative low-precision arithmetics and numerical representations. 
More specifically, this paper makes the following contributions:
\begin{itemize}
    \item Proposal of a new 12-bit arithmetic combining floating-point with stochastic rounding (Section \ref{section:floating_point}), with negligible degradation with respect to a 32-bit floating-point baseline when training neural networks considering the CIFAR-10 input set.
    \item Proposal of the context float approach, adding a scaling factor to the 12-bit floating-point arithmetic (Section \ref{section:context}). The proposed representation behaves as a regularization approach that increases the range of representable values and thus enhances the accuracy of the 32-bit baseline by 2.42\%.
    \item Proposal of the Power-of-Two neural network (Section \ref{section:poweroftwo}), a simplified model with outputs and gradients constrained to power-of-two values just using 7 bits. Due to its characteristics, the Power-of-Two neural network is able to replace costly operations such as multiplication and divisions by other hardware friendly operations such as shifts during the training face and consequently, drastically reduce the training time and the resource consumption of the models with minimal degradation.
\end{itemize}

\section{Experimental setup}
\subsection{Simulation framework}
All the experimental evaluation in this paper is done using the deep-learning framework Caffe\footnote{Available at http://caffe.berkeleyvision.org/}. To evaluate the performance of the low-precision arithmetics and representations, we constrain the values of the network model down to 12 bits. Specifically, the network parameters and intermediate values reduced to 12 bits are weights, biases, outputs, weight updates, biases updates, and gradients. However, the 12-bit representation is simulated using Caffe's double precision floating-point implementation, ensuring that the network parameters stored in higher-precision registers are always constrained to 12-bits. 

The result of arithmetic operations between two already formatted 12-bit values could lead to a non-representable value. Therefore, in order to convert the result to a 12-bit value, we saturate the value if exceeds the largest magnitude of the representation, and we make use of the stochastic rounding algorithm, which has shown great performance in previous studies \cite{fixed1,fixed2} along with 64-bit precision registers, in which we store the numerical value to be rounded:

\begin{equation}
        StochasticRound(x)=
        \begin{cases}
            \floor{x} + \epsilon &  w.p. \hspace{0.2cm} \frac{x - \floor{x}}{\epsilon}  \\
            \floor{x} & w.p. \hspace{0.2cm} 1 - \frac{x - \floor{x}}{\epsilon} \\
        \end{cases}
\end{equation}
\vspace{0.5cm}

\noindent where $x$ is the number to be rounded, $\floor{x}$ is the closest 12-bit value smaller than $x$ and $\epsilon$ is the smallest representation in terms of magnitude by the 12-bit format. When utilizing stochastic rounding, the value $x$ has more probability to be rounded to the closest 12-bit value although there also exists a smaller probability to be rounded to the second closest 12-bit value thus, preserving the information at least statistically.

One of the most executed operations in neural network training is the dot product: $A \cdot B = \sum_{i=0}^{n}a_i \cdot b_i$ where $A$ and $B$ are vectors such that each component is represented in a 12-bit format. 
In this study, when performing the dot product, the result of each multiplication $a_i \cdot b_i$ is accumulated in a higher precision variable of 64 bits of length. Only at the end of the dot product operation, the stochastic rounding or saturation methods are applied to the result.

\subsection{CNN model}
To test the performance of the different data representations, we consider a widely used image classification benchmark: the CIFAR-10 dataset\footnote{Available at https://www.cs.toronto.edu/$\sim$kriz/cifar.html}. We construct a Convolutional Neural Network (CNN) similar to the topology proposed in \cite{fixed2}. 
The CNN is made of 3 Convolutional layers followed by their corresponding Max Pooling layers and a Fully connected layer of 1000 units with dropout probability of 0,4 which is then connected to a 10-way softmax Output layer for classification. 
The first two convolutional layers consist of 32 Kernels with 5x5 dimensions and the third convolutional layer consists of 64 Kernels with 5x5 dimensions. 
All convolutional layers have stride=1 and padding=2. 
The pooling layers have dimensions 3x3 with stride=2. 
The activation function in the convolutional layers and the fully connected layer is ReLU and we define Cross-Entropy as a cost function of the model. 
We employ Stochastic Gradient Descend as a minimiser for the model with a fixed learning rate of 0.001, a momentum of 0.9 to speed up the convergence, a weight decay of 0.004 in all layers and a batch size of 100 images during the 40 epochs of training.

\section{12-bit Fixed-Point} \label{fixed_point}
The fixed-point arithmetic stands out for being fast, efficient and hardware friendly. 
By virtue of its characteristics, the fixed-point arithmetic is widely used for inference and training in neural networks. 
\cite{fixed2} demonstrate that deep networks can be trained with just 16-bit wide fixed-point numbers when stochastic rounding is used. In this section we use the same generalised fixed-point representation $fixed[I,F]$, where $I$ yields for the number of bits for the integer part of the number and $F$ the number of bits for the fractional part of the number in two's complement. 
Defining $\epsilon$ as the smallest magnitude representable, $2^{-F}$ in the case of the fixed-point, the range of representation keeps closed between $[-2^{I-1},2^{I-1}-\epsilon]$. 
After studying the magnitude of the network parameters, when training the CNN with 12-bit fixed-point arithmetic, we assign the format $fixed[0,12]$ to all the values of the network, albeit the outputs of the neurons take the form $fixed[6,6]$.

\subsection{Accuracy results for 12-bit fixed-point}

As seen in the Table \ref{table:accuracy_fixed}, 12-bit fixed-point arithmetic is not enough to train the network, stalling the accuracy that is obtained at 32$\%$. 
The incapacity of the 12-bit fixed-point for training resides on the fact that 12 bits of fraction are not sufficient for some parameters of the network, even when using stochastic rounding. 
The gradients of the network obtained from the back-propagation stage have small magnitudes and seem to diminish its magnitude slowly epoch after epoch when the network trains. 
As a result, when constraining weight updates and gradients to $fixed[0,12]$ format, a substantial amount are rounded to 0 and thus halting the network learning.

Instead of placing the point of the fixed-point representation to 0, the representation can also be scaled using a global scaling factor if the overall values of the network are significantly small. 
Enhanced results are obtained if the previous 12-bit fixed-point representation is globally scaled by $2^{-4}$ during the training phase, achieving a mean accuracy of 63\% (see Table \ref{table:accuracy_fixed}). 
Despite improving the results, the model is still far from the baseline performance.

\begin{table}[]
\small
 \centering
 \renewcommand{\arraystretch}{0.5}
    \begin{tabular}{l c c c c}
     \toprule
     \textbf{Representation } \quad & \textbf{ Accuracy } \quad & \textbf{ Accuracy no rounding } \quad & \textbf{ Epoch $\geq$ 70$\%$ } \\
     \midrule
     \midrule
     \textbf{32 bits:} &&&\\
     \midrule
     Floating-point & - & 75$,$60$\%$ $\rpm$ 0,4  & 4$,$8 epochs     \\
     \midrule
     \textbf{12 bits:} &&&\\
     \midrule
     \rowcolor{Gray}
     Fixed-point    & 32,10$\%$ $\rpm$ 1,6   & 10$\%$   & -    \\
     \rowcolor{Gray}
     Scaled Fixed-Point & 63,03$\%$ $\rpm$ 0,3   & 10$\%$   & - \\
     \bottomrule\\
    \end{tabular}
    \caption{Results of the model trained with the 12-bit fixed-point formats. The table shows the mean \textbf{Accuracy} employing the stochastic rounding algorithm, the \textbf{Accuracy no rounding} refers to the model mean accuracy when not applying any rounding algorithm and last \textbf{Epochs $\geq$ 70\%} are the epochs taken to reach at least 70\% accuracy.}
    \label{table:accuracy_fixed}
\end{table}

\section{12-bit Floating-Point} \label{section:floating_point}
As observed in the previous section, the 12-bit fixed-point really suffers the bit-width limitations. Moreover, differences in magnitudes between parameters or layers of the model requires a previous study of the scenario and an adjustment of the fixed-point format. Adjusting the format in a trial and error methodology is costly and it is something that can hindrance the utilisation of the fixed-point in low-precision scenarios.

For all of the previous reasons, we propose and evaluate the use 12-bit floating-point instead of 12-bit fixed-point. The generalised representation $float[E,M]$ is used to denote a floating-point format of $E$ bits of exponent, $M$ bits of mantissa and 1 implicit bit for the sign. We consider the format $float[5,6]$ for training the CNN. 

\subsection{Accuracy results for 12-bit floating-point}
As shown in Table \ref{table:accuracy_float}, compared to the 32-bit floating-point baseline, the 12-bit floating-point representation with stochastic rounding suffers almost no degradation. It is clearly noticed the advantage of the exponent in the floating-point representations since it allows a wider representation and more precision than the fixed-point counterpart (see Figures \ref{fig:gradients}  and \ref{fig:outputs}). Besides, unlike fixed-point, it is not required to adjust the format of the 12-bit floating-point to the network. It is also important to remark from these results the importance of using stochastic rounding; otherwise the 12-bit floating-point representation would not be able to train the network.

\begin{table}[]
\small
 \centering
  \renewcommand{\arraystretch}{0.5}
    \begin{tabular}{l c c c c}
     \toprule
     \textbf{Representation } & \textbf{ Accuracy } & \textbf{ Accuracy no rounding } & \textbf{ Epochs to $\geq$ 70\% } \\
     \midrule
     \midrule
     \textbf{32 bits:} &&&\\
     \midrule
     Floating-point & - & 75,60$\%$ $\rpm$ 0,4  & 4,8 epochs     \\
     \midrule
     \textbf{12 bits:} &&&\\
     \midrule
     Fixed-point    & 32,10$\%$ $\rpm$ 1,6   & 10\%   & -    \\
     Scaled Fixed-Point    & 63,03\% $\rpm$ 0,3   & 10\%   & - \\
    \rowcolor{Gray}
     Floating-point & 74,20$\%$ $\rpm$0,4    & 10$\%$   & 5,7    \\
     \bottomrule\\
    \end{tabular}
    \caption{Results of the model trained with the $float[5,6]$ format.}
    \label{table:accuracy_float}
\end{table}

{
\begin{figure}[]
\begin{subfigure}{.49\textwidth}
  \includegraphics[width=\textwidth,height=40mm]{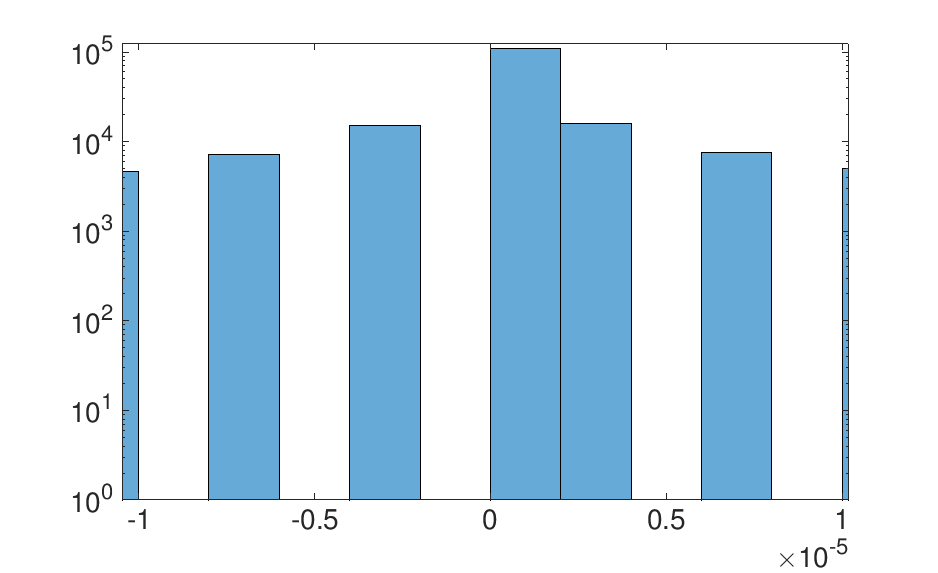}
  \centering
  \caption{12-bit fixed-point}
  \label{fig:sub1}
\end{subfigure}%
\hspace{.07\textwidth}
\begin{subfigure}{.49\textwidth}
  \centering
  \includegraphics[width=\textwidth,height=40mm]{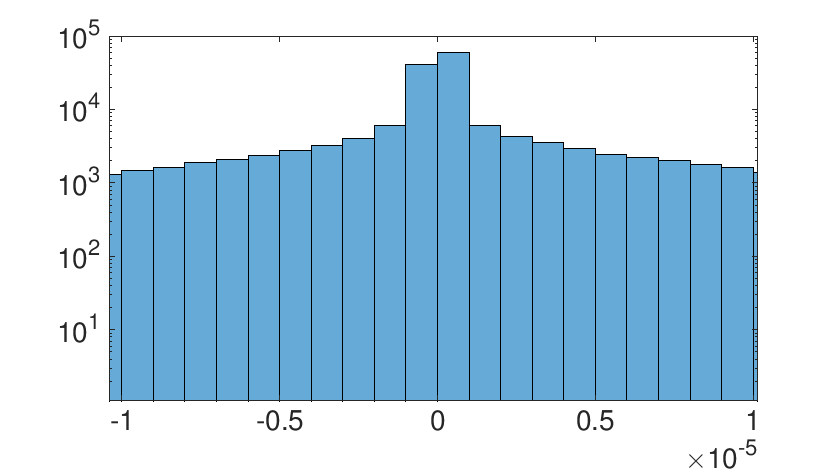}
  \caption{12-bit floating-point}
  \label{fig:sub2}
\end{subfigure}
\caption{\small Gradient values in the fully connected layer with respect to the cost function.}
\label{fig:gradients}
\end{figure}
}

{
\begin{figure}[]
\begin{subfigure}{.49\textwidth}
  \includegraphics[width=\textwidth ,height=40mm]{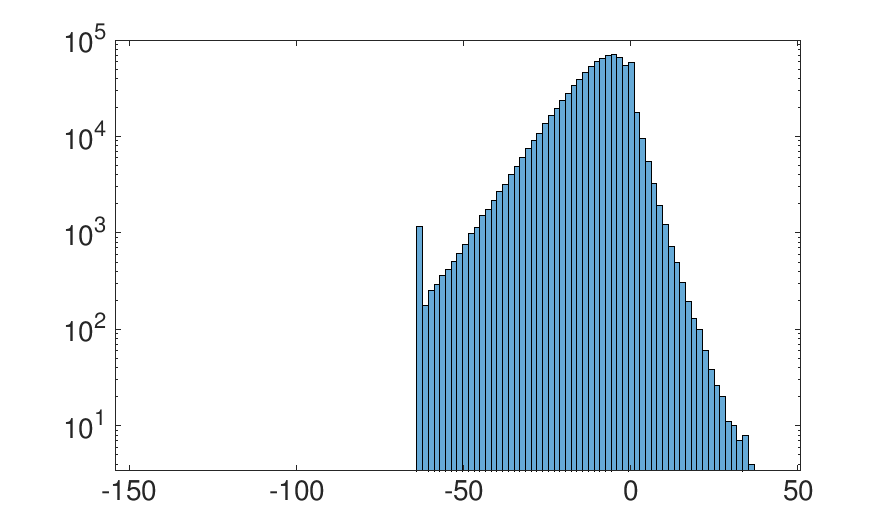}
  \centering
  \caption{12-bit fixed-point}
  \label{fig:sub1}
\end{subfigure}%
\hspace{.07\textwidth}
\begin{subfigure}{.49\textwidth}
  \centering
  \includegraphics[width=\textwidth,height=40mm]{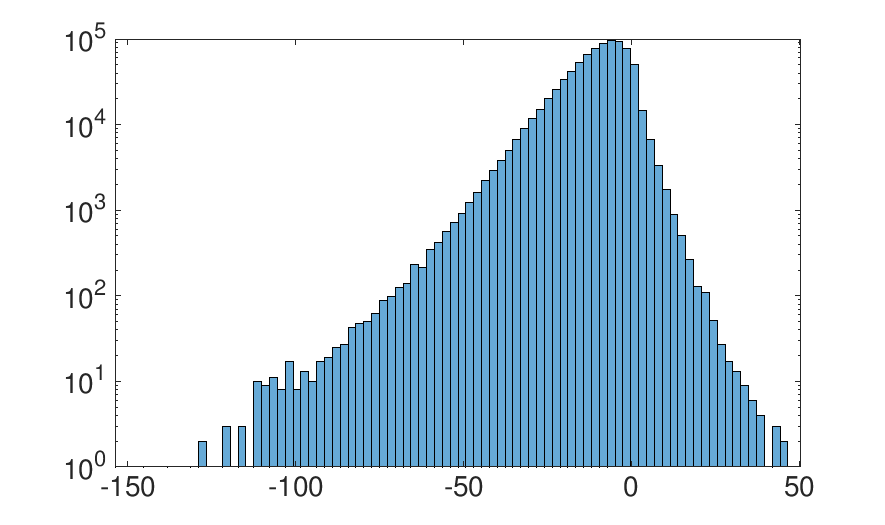}
  \centering
  \caption{12-bit floating-point}
  \label{fig:sub2}
\end{subfigure}
\caption{\small Comparing the 12-bit fixed-point and floating-point formats with a set of output values obtained from the 2nd convolutional layer of the network.}
\label{fig:outputs}
\end{figure}
}

Within the network, it is common to observe learning parameters differing in several orders of magnitude with respect to their corresponding gradients or updates.
As a consequence, the 12-bit fixed-point representation is unable to train the CNN.
Furthermore, the 12-bit floating-point approach, despite having the 5-bit exponent and thus a larger range of magnitude representation, it is unable to reach the low magnitudes that gradients and weight updates may have (see Figure \ref{fig:distributions}), slowing down the training.

\section{Context Representation} \label{section:context}

The aim of the context representation is to have the capacity of representing the network parameters independently of its magnitude with a few bits. 
We consider two variants of the context representation: i) context fixed-point and ii) context floating-point. 
Both alternatives are fixed-point/floating-point representations scaled by a local scaling factor shared among a group of parameters of the network, that we call context. 
In this work, contexts are defined by grouping parameters of the same type and the same layer: weight and biases, parameter updates, outputs and gradients. 
The scale factor is determined by computing the average of all values belonging to a certain context. 

\begin{equation}
    \begin{split}
        exponent & = \frac{1}{N}*\sum_{i=0}^{N}log2(|X_i|)\\
        ScaleFactor & = 2^{exponent}   
    \end{split}
\end{equation}

We define the context fixed-point format as $context$-$fixed[I,F]$ where $I$ corresponds to the number of bits to represent magnitudes greater than the scaling factor and $F$ to the bits to represent smaller magnitudes. 
For the evaluation of the context fixed-point we use 12 bits $context$-$fixed[6,6]$. 
On the other hand, we use the notation $context$-$float[E,M]$  to refer to a context floating-point that uses one bit for the sign, a scaled exponent (with $E$ bits) and a mantissa (with $M$ bits). 
For the evaluation of the context floating-point we use 12 bits $context$-$float[4,7]$ format (observe the bit subtracted from the exponent and placed in the mantissa for precision improvement). 
When assigning 4 bits to the exponent, exponent values higher than $0000_2$, in two's complement, express magnitudes higher than the scaling factor and lower exponents express magnitudes lower than the scaling factor. 

{
\begin{figure}[]
\centering
\includegraphics[width=\textwidth, height=6cm]{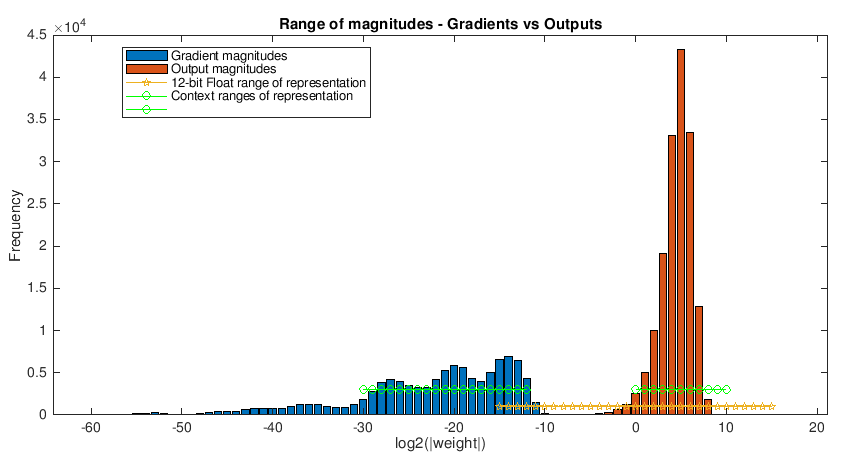}
\caption{\small{The x-axis correspond to the $log2$ of the absolute values of the network, and the y-axis, is the frequency of appearance of the values. In \textbf{blue} the gradient values of the fully connected layer for a certain iteration $i$. In \textbf{red} the output values of the last convolutional layer at the same iteration $i$. The range of representation of the $float[5,6]$ and a possible range of the context representation are displayed with \textbf{orange} and \textbf{green} lines.}}
\label{fig:distributions}
\end{figure}
}

Previous work has also proposed a similar approach as an extension of the typical fixed-point arithmetic for training neural networks.
For example, \cite{dynamic_fixedpoint} proposes to update the scaling factor depending on the number of overflows observed for a certain period during training.
Recently ~\cite{flexpoint} predicts the optimal scaling factor by means of a dynamic algorithm.
In this section, we analyse the context fixed-point representation using fewer bits than these previous works and extend it to our context floating-point proposal.

\subsection{Accuracy results for context fixed- and floating-point}
As shown in Table \ref{table:results_context}, the proposed $context$-$fixed[6,6]$ format completely introduces no degradation in accuracy; surprisingly, the $context$-$float[4,7]$ with stochastic rounding, with just 12 bits of representation, surpasses the 32-bit floating-point model with stochastic rounding. In addition, $context$-$float[4,7]$ achieves decent results even without using stochastic rounding. 
\begin{table}[H]
\small
 \centering
   \renewcommand{\arraystretch}{0.5}
    \begin{tabular}{l c c c c}
     \toprule
     \textbf{Representation } & \textbf{ Accuracy } & \textbf{ Accuracy no rounding } & \textbf{ Epochs to $\geq$ 70\% } \\
     \midrule
     \midrule
     \textbf{32 bits:} &&&\\
     \midrule
     Floating-point & $-$ & 75,60\% $\rpm$ 0,4  & 4,8     \\
     \midrule
     \textbf{12 bits:} &&&\\
     \midrule
     Fixed-point    & 32,10$\%$ $\rpm$ 1,6   & 10$\%$   & $-$     \\
     Scaled fixed-point    & 63,03$\%$ $\rpm$ 0,3   & 10$\%$   & $-$ \\
     Floating-point & 74,20$\%$ $\rpm$ 0,4    & 10$\%$   & 5,7     \\
          \rowcolor{Gray}
     Context-fixed & 76,32$\%$ $\rpm$ 0,5    & 10$\%$   & 5      \\
          \rowcolor{Gray}
     Context-float & 78,02\% $\rpm$ 0,3    & 71,88\% $\rpm$ 0,4   &  5   \\
     \bottomrule
    \end{tabular}
    \caption{Training results for the \textit{Context-fixed[6,6]} and the \textit{Context-float[4,7]} representations.}
    \label{table:results_context}
\end{table}

We attribute the high accuracy of the $context$-$float[4,7]$ representation to the following two reasons.  
The first reason is trivial, the range of representation and precision increase; this factor allowed the network to train with no rounding algorithm. 
The second and not so obvious reason is that the trimming of bits in an intelligent manner may be a valid regularization technique. 
In the specific case of the weights, they have a tendency to follow a Gaussian distribution which gets wider as the network learns, distinguishing relevant features and thus increasing its assigned weight or decreasing the weight otherwise (Figure \ref{fig:context_weights}). 
Allocating fewer bits in the scaled exponent limits the width of the distribution and constraints the network from learning too much (Figure \ref{fig:context_training}).

{
\begin{figure}[]
\centering
\includegraphics[width=8cm, height=5.5cm]{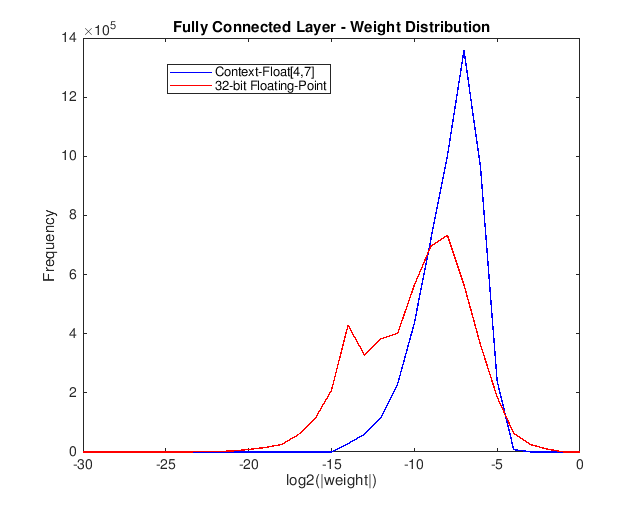}
\caption{\small{In \textbf{red} the distribution of weights of the fully connected layer when training the network with a 32-bit floating point representation. In \textbf{blue} the distribution of weights of the same layer in the same iteration, when training the network with the $context$-$float[4,7]$ representation.}}
\label{fig:context_weights}
\end{figure}
}

{
\begin{figure}[]
\centering
\includegraphics[width=\textwidth, height=8cm]{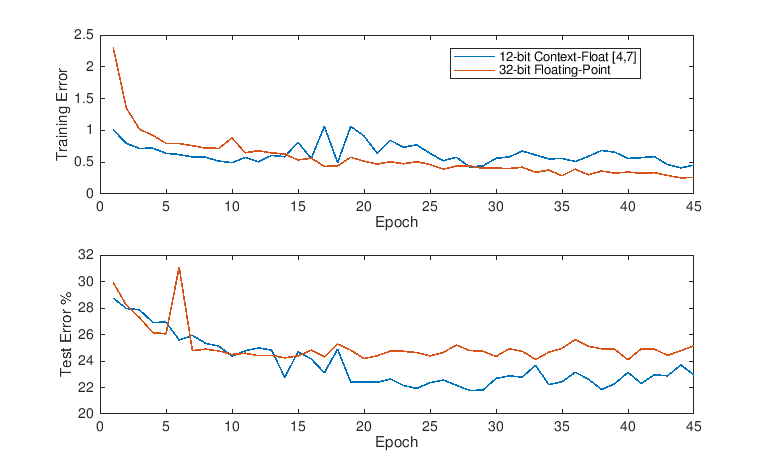}
\caption{\small{Training evolution of the CNN.}}
\label{fig:context_training}
\end{figure}
}

\section{Power-of-Two Neural Network} \label{section:poweroftwo}
The power-of-two neural network combines the advantages of the efficient fixed-point arithmetic and the floating-point representation. 
When training the network with 12 bits and fixed-point representation many of its key parameters are not representable and thus rounded to 0, which stalls the network training. 
In order to represent these small magnitudes, we use a variant of the already evaluated $float[5,6]$: the $float[6,0]$ format. With this format, only the exponent and sign bits are used to represent the network gradients and the output values. 
This scheme uses fixed-point $fixed[0,12]$ representation for the rest of values (weights, biases, weight updates and bias updates).

The purpose of constraining gradients and outputs to power-of-two values is not only to reduce the memory requirements of the model. 
Since the network trains with fixed-point arithmetic and outputs and gradients hold a power-of-two value, the costly floating-point operations of multiplication and division that are needed when using 32-bit arithmetic can now be replaced by simple shifts between: a) weights in $fixed[0,12]$ and inputs (outputs of a previous layer) in $float[6,0]$ when performing the forward propagation, and b) weights and gradients in $float[6,0]$ in the backward propagation.

\subsubsection{a) Forward propagation\\\\}

In the forward propagation, the output for each neuron $Out$ is computed by applying an activation function $f$ over the potential of the neuron $P$. The potential of the neurons is determined by a costly dot product between the outputs of the neurons of the previous layer $X$ in $float[6,0]$ format, and the weights associated $W$ in $fixed[0,12]$ format:  

  \begin{equation}
     Out = f(P)\\
  \end{equation}
  \begin{equation}
     P = \sum_{j=1}^{n} X_jW_{j} \approx \sum_{j=1}^{n} 2^yW_{j} = \sum_{j=1}^{n} W_{j} << y\\ 
  \end{equation}

With ReLU as activation function $f$, max pooling in the convolutional layers and the outputs constrained in the form $2^y$, multiplications and divisions in the forward propagation of the CNN are replaced by simple shifts. 
While the dot product is being computed, the intermediate values of $P$ and $Out$ are stored in higher precision variables. 
Moreover, $Out$ has to be formatted to $float[6,0]$ in order to avoid multiplications and divisions in the following layers of the network (Figure \ref{fig:poweroftwo_sample}).\\

{
\begin{figure}[]
\centering
\includegraphics[width=7cm]{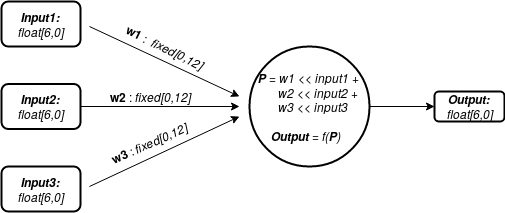}
\caption{\small{Simulation of the forward-propagation algorithm in the Power-of-Two model with 1 neuron and 3 connections. Each parameter displays its corresponding format in the model.}}
\label{fig:poweroftwo_sample}
\end{figure}
}

\subsubsection{b) Backward propagation\\\\}

The network learns by updating each of the learning parameters so that they cause the actual output to be closer the target output, thereby minimizing the error for each output neuron and the network as a whole. 
The updates of all learning parameters $w$ are done by modifying it with its gradient towards the local minimum of the cost function. 
The gradient of a learning parameter can be computed by performing a partial derivative as follows:
\begin{equation}\label{eq:chain}
    \frac{\partial Cost}{\partial w} = \frac{\partial Cost}{\partial f(x)} \times \frac{\partial f(x)}{\partial P} \times \frac{\partial P}{\partial w}
\end{equation}

Where $f(x)$ is the activation function and $P$ the potential of the neuron. 
$\frac{\partial f(x)}{\partial P}=1$ with ReLU activation functions. 
The gradient of the potential of the neuron with respect to the cost function is constrained to $float[6,0]$, a power of two value:
\begin{equation}
    \frac{\partial Cost}{\partial P} = \frac{\partial Cost}{\partial f(x)} \times \frac{\partial f(x)}{\partial P} \approx 2^x
\end{equation}

Therefore, the gradient of a learning parameter required to compute the parameter update (Equation \ref{eq:weight_update}) and the propagation of gradients to other connected hidden neurons, also known as error propagation (Equation \ref{eq:backprop}) become now:

\begin{equation} \label{eq:weight_update}
    \frac{\partial Cost}{\partial w} = \frac{\partial Cost}{\partial f(x)} \times \frac{\partial f(x)}{\partial P} \times \frac{\partial P}{\partial w} \approx 2^x \times \frac{\partial P}{\partial w}
\end{equation}

\begin{equation} \label{eq:backprop}
     \frac{\partial Cost}{\partial f(x)_i^{l-1}} = \sum_{j=1}^{n}  \frac{\partial Cost}{\partial P_j^{l}} w_{ji}^{l} \approx \sum_{j=1}^{n}  2^y w_{ji}^{l} \\ 
\end{equation}

Where $l$ is a layer in the network, $i,j$ neurons and $w_{ji}$ the weight of the connection between neurons $i,j$. 

Although not realized in this study, in order to replace multiplications and divisions by shifts in weight update computations, the learning rate ($\alpha$), momentum ($\mu$) and other hyperparameters could also be expressed as a power of two:
  \begin{equation}
     \text{New }w = w + \mu\Delta w^{-1} - \alpha \Delta w \hspace{0.1cm}\approx w + \hspace{0.1cm} 2^y\Delta w^{-1} - 2^z \Delta w
\end{equation}
where $w$ is a learning parameter and $\Delta w$ is the gradient of the learning parameter with respect to the cost function.

The Power-of-Two neural network is based on previous works~\cite{ternary,binary} where the reduction in resource consumption is pushed to the limit when training networks. 
Results obtained from those studies are yet impressive although those super-simplified models are not able to train on their own and need a side high-precision model to be quantized and thus, the memory requirements are not lowered. 
The Power-of-Two proposal in this paper does reduce the memory requirements during both inference and training, has a lower quantization overhead, trains with no auxiliary regularization techniques and is able to evade all the multiplications and divisions during training and inference. 

\subsection{Accuracy results for Power-of-Two}

As shown in Table \ref{table:results_pot}, the simplified power-of-two neural network brings only a 2\% average accuracy degradation from the baseline model, the 32-bit floating-point while achieving drastic reductions in training time, memory requirements and energy consumption, as it is shown in section~\ref{sec:time}.
If the 2\% degradation of the power-of-two network is considered excessive, it would be possible to use it to accelerate the training up to a certain accuracy and later switch to a more accurate floating-point arithmetic, for instance.

\begin{table}[H]
\small
 \centering
 \renewcommand{\arraystretch}{0.1}
    \begin{tabular}{l c c c c}
     \toprule
     \textbf{Representation } & \textbf{ Accuracy } & \textbf{ Accuracy no rounding } & \textbf{ Epochs to $\geq$ 70\% } \\
     \midrule
     \midrule
     \textbf{32 bits:} &&&\\
     \midrule
     Floating-point & $-$ & 75,60\% $\rpm$ 0,4  & 4,8\\
     \midrule
     \textbf{12 bits:} &&&\\
     \midrule
     Fixed-point    & 32,10$\%$ $\rpm$ 1,6   & 10$\%$   & $-$\\
     Scaled fixed-point    & 63,03$\%$ $\rpm$ 0,3   & 10$\%$   & $-$\\
     Floating-point & 74,20$\%$ $\rpm$ 0,4    & 10$\%$   & 5,7\\
     Context-fixed & 76,32$\%$ $\rpm$ 0,5    & 10$\%$   & 5\\
     Context-float & 78,02\% $\rpm$ 0,3    & 71,88\% $\rpm$ 0,4   &  5\\
     \midrule
     \textbf{7 or 12 bits}\footnotemark\\
     \midrule
     \rowcolor{Gray}
     Power-Of-Two  & 73,42\% $\rpm$ 0,3    & 10\%   &  18,6\\
     \bottomrule
    \end{tabular}
    \caption{Training results for the power-of-two model.}
    \label{table:results_pot}
\end{table}

\footnotetext{7 bits for outputs and gradients, 12 bits for the rest of the network parameters.}

\section{Time Results and Memory Requirements}
\label{sec:time}
In this section, we estimate the memory requirements and training time of the different analysed arithmetics in this paper. The estimation corresponds to the CNN model until epoch 40 of the training phase. The time estimation only considers the time required to perform arithmetic computations and other intermediate operations needed to preserve the format of the data. Neither rounding algorithms overhead nor memory accesses and penalizations are considered as if an ideal neural network training was referred. In practice, we compute the total number of arithmetic operations needed to train the model until epoch 40 and the estimated cost of each arithmetic operation. The objective of the estimation is to perceive the magnitude of an improvement and not an exact measurement. The estimated cost of each arithmetic operation is obtained from an architecture based on an Intel E5649 (6-Core) processor running at 2.53 GHz.

Most of the current processors make use of vector (SIMD, single-instruction multiple-data) extensions. With these vector extensions, the central processing unit can operate on a sequence of elements in a single instruction, making use of data parallelism. In the estimation of the training time we consider the use of vector operations so, ideally, the time that the processor takes to perform a 32-bit operation is the same that a processor takes to perform 2,66 12-bit operations. With that strategy in mind, it is guaranteed that the elements in the operand vectors are adjacent in memory to benefit from the vector operations and exploit the memory hierarchy.

\begin{figure}[ht]
\centering
\begin{tikzpicture}
    \begin{axis}[
        width  = \textwidth,
        height = 5cm,
        major x tick style = transparent,
        ybar=\pgflinewidth,
        bar width=4pt,
        ymajorgrids = true,
        ylabel = {MBytes},
        x tick label style={font=,rotate=60},
        symbolic x coords={Total, ConvLayer1, ReluLayer1, MaxPool, ConvLayer2, ReluLayer2, AvePool1, ConvLayer3, AvePool2, ReluLayer3, FullyConnLayer},
        xtick = data,
        scaled y ticks = false,
        ytick distance = 2,
        enlarge x limits=0.05,
        legend cell align={left},
        ymin=0
    ]
        \addplot [color=black, fill=amber]
            coordinates {(Total,12.702256) (ConvLayer1,3.538944) (ReluLayer1, 0.393216) (MaxPool, 0.0576) (ConvLayer2, 2.9952) (ReluLayer2, 0.1152) (AvePool1, 0.016384) (ConvLayer3, 0.851968) (AvePool2, 0.004608) (ReluLayer3, 0.004608) (FullyConnLayer, 4.632)};
        
        \addplot [color=black, fill=bblue]
            coordinates {(Total,4.763346) (ConvLayer1,1.327104) (ReluLayer1,0.147456) (MaxPool, 0.0216) (ConvLayer2, 1.1232) (ReluLayer2, 0.0432) (AvePool1, 0.006144) (ConvLayer3, 0.319488) (AvePool2, 0.001728) (ReluLayer3, 0.001728) (FullyConnLayer, 1.737)};

        \addplot [color=black, fill=ggreen]
            coordinates {(Total,3.783888) (ConvLayer1,1.277952) (ReluLayer1, 0.073728) (MaxPool, 0.0108) (ConvLayer2, 0.8316) (ReluLayer2, 0.0216) (AvePool1, 0.003072) (ConvLayer3, 0.236544) (AvePool2, 0.000864) (ReluLayer3, 0.000864) (FullyConnLayer, 1.302)};

        \legend{32-bit Floating-point model , 12-bit Floating-point/Fixed-point/Context models, 12-bit Power-of-two model}
    \end{axis}
\end{tikzpicture}
\caption{\small{Memory requirements per model.}}
\label{fig:memory}
\end{figure}
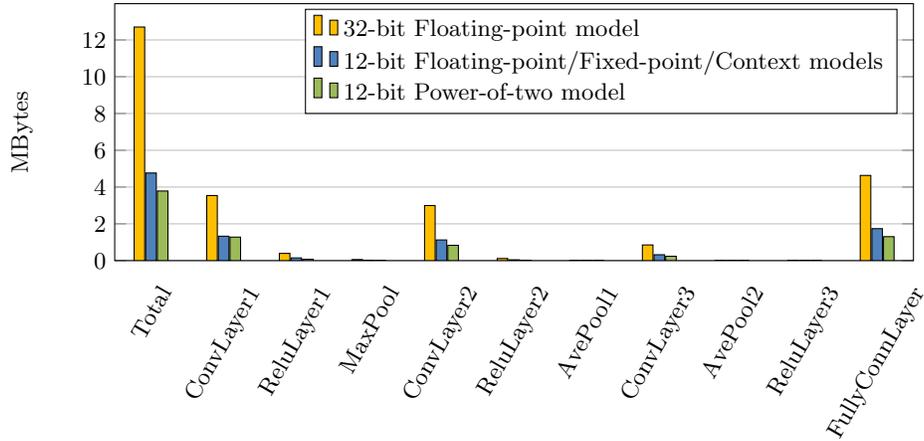

As shown in Figures \ref{fig:time} and \ref{fig:memory}, the power-of-two network reduces drastically the training time, and it is estimated to achieve an 8x speedup with respect to the 32-bit floating-point baseline. Furthermore, the real speedup may increase as the power-of-two network should reduce memory penalization, not taken into account in this study. The network trained with a $context$-$float[4,7]$ representation is much slower due to the fact that the floating-point had to be scaled previously to any operation between two parameters of different contexts. Finally the network with $context$-$fixed[6,6]$ is estimated to be as complex as the CNN with floating-point arithmetic.

\begin{figure}[ht]
\centering
\begin{tikzpicture}
    \begin{axis}[
        width  = \textwidth,
        height = 5cm,
        major x tick style = transparent,
        ybar=\pgflinewidth,
        bar width=3pt,
        ymajorgrids = true,
        ylabel = {Estimated training hours},
        x tick label style={font=,rotate=60},
        symbolic x coords={Total, ConvLayer1, ReluLayer1, MaxPool, ConvLayer2, ReluLayer2, AvePool1, ConvLayer3, AvePool2, ReluLayer3, FullyConnLayer},
        xtick = data,
        scaled y ticks = false,
        ytick distance = 0.25,
        enlarge x limits=0.05,
        legend cell align={left},
        ymin=0
    ]
        \addplot [color=black, fill=amber]
            coordinates {(Total,2.0) (ConvLayer1,0.75) (ReluLayer1, 0) (MaxPool, 0) (ConvLayer2, 0.45) (ReluLayer2, 0) (AvePool1, 0) (ConvLayer3, 0.15) (AvePool2, 0) (ReluLayer3, 0) (FullyConnLayer, 0.55)};
            
        \addplot [color=black, fill=ppurple]
            coordinates {(Total,1.291214067) (ConvLayer1,0.4846933333) (ReluLayer1, 0) (MaxPool, 0) (ConvLayer2, 0.2955) (ReluLayer2, 0) (AvePool1, 0.001789866667) (ConvLayer3, 0.08405333333) (AvePool2, 0.0004917000001) (ReluLayer3, 0) (FullyConnLayer, 0.41754)};
        
        \addplot [color=black, fill=bblue]
            coordinates {(Total,0.751287616675702) (ConvLayer1,0.281531733336696) (ReluLayer1, 0) (MaxPool, 0) (ConvLayer2, 0.170940000002209) (ReluLayer2, 0) (AvePool1, 0.000967466667022165) (ConvLayer3, 0.0486229333339617) (AvePool2, 0.000260850000054984) (ReluLayer3, 0) (FullyConnLayer, 0.244768800002385)};
            
        \addplot [color=black, fill=rred]
            coordinates {(Total,0.3739916292) (ConvLayer1,0.1394005333) (ReluLayer1, 0) (MaxPool, 0) (ConvLayer2, 0.08352000001) (ReluLayer2, 0) (AvePool1, 0.0002941333333) (ConvLayer3, 0.0237568) (AvePool2, 0.0000797625) (ReluLayer3, 0) (FullyConnLayer, 0.1247904)};
            
        \addplot [color=black, fill=ggreen]
            coordinates {(Total,0.2282776801) (ConvLayer1,0.08577260118) (ReluLayer1, 0) (MaxPool, 0) (ConvLayer2, 0.01487489605) (ReluLayer2, 0) (AvePool1, 0.001482428904) (ConvLayer3, 0.0004170103786) (AvePool2, 0.0000797625) (ReluLayer3, 0) (FullyConnLayer, 0.0721198769)};

        \legend{32-bit Floating-point, 12-bit Context-Float, 12-bit Floating-point/12 bit Context-Fixed, 12-bit Fixed-point, Power of two}
    \end{axis}
\end{tikzpicture}
\caption{\small{Training time estimation for the CNN until epoch 40.}}
\label{fig:time}
\end{figure}
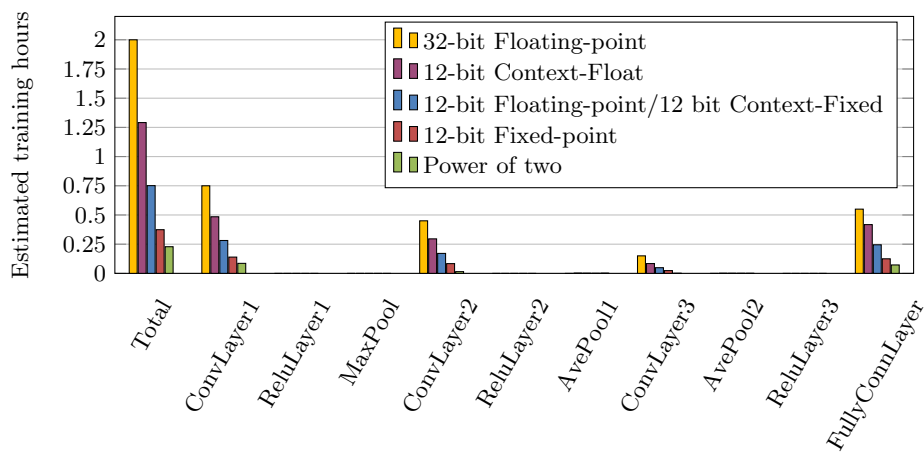

\section{Related Work}

Low precision computations for neural networks is not a recent idea since Iwata et al. \cite{float24} earlier suggested a backpropagation algorithm for neural networks with just 24 bits wide floating-points. Besides, D. Hammerstrom \cite{hammer}, in order to understand the behavior of a neural network with limited bit width computations, trained a neural network model with an 8 or 16-bit fixed-point arithmetic. All of the previous studies have been carried out successfully in fairly simple models though, in the deep-learning field, reducing the precision may be more tedious and desired. Studies \cite{vanhoucke} and \cite{gong} show that it is possible to train and perform inference in deep neural networks with low-precision fixed-point representations. Moreover, S.Gupta et al. \cite{fixed2} assert that deep-neural networks can be trained with as few as 16 bits of representation and fixed-point arithmetic with stochastic rounding, and incur little or no degradation in the model.  This paper is inspired by the study \cite{fixed2} and tries to demonstrate that fixed-point arithmetic with a limited range of representation even with stochastic rounding, it is not the best choice since it is known that the magnitudes of the parameters may vary significantly not only between networks but also between layers or parameters \cite{jud}. Studies by M. Courbariaux et al.  \cite{flexpoint} and U. Köster et al.\cite{dynamic_fixedpoint} aware of that barrier, propose a fixed-point representation where the fixed-point does not contain a global scaling factor but multiple local scaling factors referencing sets of values of the network and so, being able to extrapolate the fixed-point arithmetic to other models. Section \ref{section:context} in this paper references and analyzes the performance of the locally scaled fixed-point presented by studies \cite{flexpoint,dynamic_fixedpoint} and complements the concept with novel proposals. Finally, there has been a recent interest in pushing resource reduction to the limit and several studies have been proposed in which deep neural network's training time and resource consumption is reduced drastically. One of the studies (\cite{binary}) quantizes weights in real-time from full-precision to +1 or -1 values, replacing most of the training operations for bit-wise operations. The concept of the power-of-two neural network detailed in the Section \ref{section:poweroftwo} is also influenced by these ideas and we propose a simplified model that considerably reduces resource requirements. Unlike \cite{binary}, with the power-of-two network we are able to reduce memory requirements while training, reduce the overhead of quantization and normalization and lower the training time with little degradation in performance.

\section{Conclusions}
In this paper, we show that the commonly used and efficient fixed-point arithmetic, even with stochastic rounding, faces serious obstacles when training neural networks with limited numerical representation. Based on this initial analysis, a manifold of arithmetics have been analyzed along with stochastic rounding as an alternative. First off a 12-bit float representation has been explored with acceptable results, showing its ability to train the network with little accuracy degradation with respect to the 32-bit floating-point network baseline. Thereafter, we have experimented with locally scaled fixed-point and floating-point configurations with positive results. The 12-bit scaled fixed-point completely avoids degradation when reducing the bit width and on the other hand, the scaled floating-point enhances the accuracy results of the baseline with just 12 bits of representation; in contrast to the other low-precision models seen previously, the scaled floating-point is able to learn with no rounding algorithm. Last but not least, a simplified fixed-point model is presented, which avoids multiplications and divisions by constraining key parameters with powers of two, reducing drastically training time, energy consumed and memory with negligible degradation in accuracy results.

We believe that in the future, the success of the utilisation of neural networks in real-life problems will be dependent on models, arithmetics and representations conscious on the underlying hardware architecture. Moreover, the hardware should be designed taking advantage of the capabilities of the network to tolerate noise; for example having the possibility to perform low-precision computations and combine low-precision and higher precision arithmetics at run-time in order to minimise the resources when needed without impacting accuracy.

\section*{Acknowledgments}
This work is partially supported by the Spanish Government through Programa Severo Ochoa (SEV-2015-0493), by the Spanish Ministry of Science and Technology through TIN2015-65316-P project and by the Generalitat de Catalunya (contracts 2017-SGR-1414 and 2017-SGR-1328).

\bibliographystyle{IEEEtran}
\bibliography{./ms}

\end{document}